\documentclass[sigconf]{acmart}
\AtBeginDocument{%
  }

\setcopyright{acmlicensed}
\copyrightyear{2018}
\acmYear{2018}
\acmDOI{XXXXXXX.XXXXXXX}

\acmConference[Preprint]{Make sure to enter the correct
  conference title from your rights confirmation email}{Oct,
  2025}{}

\acmISBN{978-1-4503-XXXX-X/2018/06}

\usepackage{enumitem}
\usepackage{algorithm,algpseudocode,placeins,array,makecell,multirow}

\usepackage{algorithmicx}
\usepackage{algpseudocode}   

\usepackage{amsmath, amssymb}

\usepackage{enumitem}

\usepackage{stfloats}

\setcopyright{none}
\renewcommand\footnotetextcopyrightpermission[1]{}
\newcommand{\policy}{\pi_{\theta}}
\newcommand{\refpolicy}{\pi_{\mathrm{ref}}}
\newcommand{\oldpolicy}{\pi_{\theta_{\mathrm{old}}}}
\newcommand{\verification}{\mathcal{V}}
\newcommand{\history}{\mathcal{H}}
\newcommand{\reals}{\mathbb{R}}
\newcommand{\expect}{\mathbb{E}}
\newcommand{\kl}{D_{\mathrm{KL}}}
\newcommand{\rewardmodel}{R_\phi}
\newcommand{\queryspace}{\mathcal{Q}}
\newcommand{\itemspace}{\mathcal{I}}
\newcommand{\outputspace}{\mathcal{O}}
\newcommand{\stepoutputspace}{\mathcal{Y}}
\newcommand{\dataset}{\mathcal{D}}
\newcommand{\indicator}{\mathbb{I}}

\begin{document}

\title{SHE: Stepwise Hybrid Examination Reinforcement Learning Framework for E-commerce Search Relevance}

\author{Pengkun Jiao}
\authornote{Equal contribution.}
\email{pkjiao23@m.fudan.edu.cn}
\affiliation{%
  \institution{Fudan University}
  \city{Shanghai}
  \country{China}
}

\author{Yiming Jin}
\authornotemark[1]
\authornote{Corresponding author.}
\email{imayking13@gmail.com}
\affiliation{%
  \institution{Taobao \& Tmall Group of Alibaba}
  \city{Hangzhou}
  \country{China}}

\author{Jianhui Yang}
\email{yangjh23@mails.tsinghua.edu.cn	}
\affiliation{%
 \institution{Tsinghua University}
 \city{Beijing}
 \country{China}}

\author{Chenhe Dong}
\email{dongchenhe.dch@alibaba-inc.com}
\affiliation{%
  \institution{Taobao \& Tmall Group of Alibaba}
  \city{Hangzhou}
  \country{China}
}

\author{Zerui Huang}
\email{huangzerui.hzr@taobao.com}
\affiliation{%
  \institution{Taobao \& Tmall Group of Alibaba}
  \city{Hangzhou}
  \country{China}}

\author{Shaowei Yao}
\email{yaoshaowei@taobao.com}
\affiliation{%
  \institution{Taobao \& Tmall Group of Alibaba}
  \city{Hangzhou}
  \country{China}
}


\author{Dan Ou}
\email{oudan.od@taobao.com}
\affiliation{%
  \institution{Taobao \& Tmall Group of Alibaba}
  \city{Hangzhou}
  \country{China}}

\author{Haihong Tang}
\email{piaoxue@taobao.com}
\affiliation{%
  \institution{Taobao \& Tmall Group of Alibaba}
  \city{Hangzhou}
  \country{China}}

\renewcommand{\shortauthors}{Jiao et al.}

\begin{abstract}
Query-product relevance prediction is a foundational technology in e-commerce search engines and has become increasingly important in AI-driven e-commerce. The recent emergence of large language models (LLMs), particularly their chain-of-thought (CoT) reasoning capabilities, offers promising opportunities for developing relevance systems that are both more interpretable and more robust. However, existing training paradigms have notable limitations: SFT and DPO suffer from poor generalization on long-tail queries and from a lack of fine-grained, stepwise supervision to enforce rule-aligned reasoning.
In contrast, reinforcement learning with verification rewards (RLVR) suffers from sparse feedback, which provides insufficient signal to correct erroneous intermediate steps, thereby undermining logical consistency and limiting performance in complex inference scenarios.

To address these challenges, we introduce the \textbf{S}tepwise \textbf{H}ybrid \textbf{E}xamination Reinforcement Learning framework for search relevance (SHE). At its core is Stepwise Reward Policy Optimization (SRPO), a reinforcement learning algorithm that leverages step-level rewards generated by a hybrid of a high-quality generative stepwise reward model and a human-annotated offline verifier, prioritizing learning from critical correct and incorrect reasoning steps. 
To bolster robustness and generalization, SHE further integrates a dual-strategy optimization: diversified data filtering, which broadens the exploration of reasoning trajectories to preempt policy entropy collapse, and a multi-stage curriculum learning protocol that systematically orchestrates progressive capability acquisition. 
Extensive experiments on real-world search benchmarks show that SHE improves both reasoning quality and relevance-prediction accuracy in large-scale e-commerce settings, outperforming SFT, DPO, GRPO, and other baselines, while also enhancing interpretability and robustness.
\end{abstract}



\begin{teaserfigure}
    \centering
    \includegraphics[width=1\linewidth]{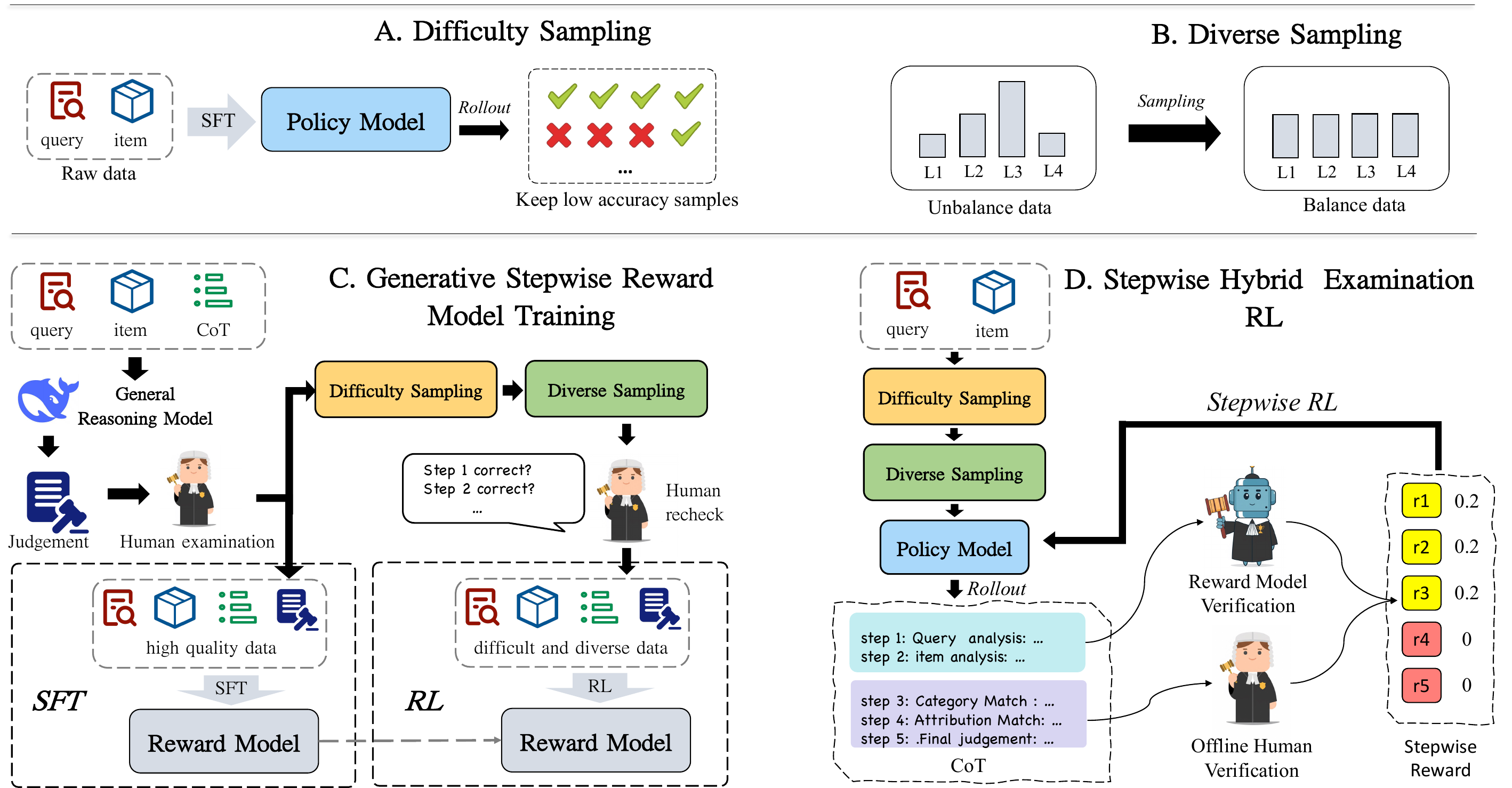}
    \caption{SHE integrates several key techniques for advanced reinforcement learning, including: (A) Difficulty Sampling, (B) Diverse Sampling, (C) Generative Stepwise Reward Model Training, and (D) Stepwise Hybrid Examination RL framework.}
    \label{fig:main}
\end{teaserfigure}

\received{20 February 2007}
\received[revised]{12 March 2009}
\received[accepted]{5 June 2009}

\maketitle

\section{Introduction}

E-commerce search relevance~\cite{karmaker2017application,yao2022reprbert,dong2025taosr1}, the task of evaluating the semantic correspondence between a user's query and a product, is a cornerstone technology for huge platforms like Taobao and Amazon. Its foundational role has become even more critical with the widespread integration of generative AI~\cite{devlin2019bert,achiam2023gpt4} and conversational interfaces in modern e-commerce search.

Traditionally, search relevance has been framed as a classification problem. However, such discriminative models are often criticized for their "black-box" nature, offering limited insight into their decision-making process. The advent of Large Language Models (LLMs)\cite{achiam2023gpt4,yang2025qwen3} has introduced a new paradigm: leveraging their capacity for explicit, step-by-step reasoning via Chain-of-Thought (CoT)~\cite{wei2022cot} to perform relevance judgments. This paradigm shifts the focus from scalar predictions to structured reasoning processes. While this explicit format offers a promising path toward transparency, recent studies suggest that the generation of reasoning steps requires rigorous alignment to ensure they faithfully reflect the model's decision logic and robustness~\cite{turpin2023llm_don_always}. Therefore, optimizing the quality and consistency of these reasoning paths becomes essential for building a truly trustworthy search relevance system.


Despite this promise, effectively training these models for search relevance presents significant challenges. Prevailing methods like Supervised Fine-Tuning (SFT) and Direct Preference Optimization (DPO)~\cite{rafailov2023dpo} often lead to models that overfit the training data, exhibiting poor generalization on complex or out-of-distribution queries. Meanwhile, Reinforcement Learning with Verifiable Rewards (RLVR)~\cite{shao2024grpo} is constrained by sparse rewards, where feedback is delivered only upon the correctness of the final output. This inefficient credit assignment forces the model to navigate a vast search space with minimal guidance, often hindering effective learning and promoting reward hacking~\cite{skalse2022reward_game}.

To address these limitations, we propose \textbf{SHE}: a Stepwise Hybrid Examination Reinforcement Learning Framework for E-commerce Search Relevance. Our framework makes the following key contributions:
\begin{itemize}[leftmargin=10pt]
    \item We design a data-centric training strategy that enhances efficiency. By constructing a highly diverse dataset and employing \textbf{offline rejection sampling}, we filter out uninformative training instances (where all reasoning paths are uniformly correct or incorrect) to accelerate convergence.
    \item We implement a dynamic training paradigm that combines \textbf{hard-sample mining} with \textbf{curriculum learning}. This allows the training curriculum to adapt to the model's evolving capabilities, progressively focusing on more challenging samples to systematically improve its reasoning prowess.
    \item We introduce a \textbf{Stepwise Hybrid Reward} mechanism that provides dense, step-specific feedback by combining scores from a trained generative stepwise reward model with verifiable ground-truth signals. This is coupled with a \textbf{Stepwise Reward Policy Optimization (SRPO)} that properly assigns credit to each reasoning step, mitigating the sparse reward problem.
\end{itemize}

\noindent Extensive offline and online evaluations on e-commerce search-relevance datasets demonstrate the effectiveness of SHE and advance the reasoning capabilities of LLMs.

\section{Related Works}
\subsection{Search Relevance}
Relevance modeling is a core task in information retrieval (IR), aiming to capture the semantic relationship between queries and items. The field has evolved over decades. Early approaches, such as BM25~\cite{robertson2009BM25} and TF-IDF~\cite{singhal2001modern_ir_tfidf}, relied on manually engineered features to estimate relevance. While effective in early IR systems, these feature-based methods suffered from poor cross-domain generalization and required substantial human effort for feature design. 

Subsequently, the advent of deep learning brought representation-based models~\cite{hambarde2023IR_advances_and_beyond_repreasentation_based}, which encode queries and items into dense vectors and compute their semantic similarity. These methods eased the burden of manual feature engineering, but their limited access to broad world knowledge hindered performance on complex or nuanced cases. The introduction of BERT~\cite{devlin2019bert} and the Transformer architecture~\cite{vaswani2017attention} marked a major paradigm shift: relevance modeling via fine-tuned pre-trained language models~\cite{devlin2019bert,achiam2023gpt4}. This approach captures richer semantic relationships, unifies task formulation, and enables end-to-end optimization with minimal feature engineering. More recently, decoder-only large language models (LLMs) have demonstrated exceptional capabilities in both natural language understanding and generation, blurring the line between upstream and downstream NLP tasks. This has led to LLM-based relevance modeling frameworks — such as LREF~\cite{tang2025lref}, ProRBP~\cite{chen2024proRBP}, and the method by Mehrdad et al.~\cite{mehrdad2024llm4relevance} — which employ techniques like supervised fine-tuning (SFT) and direct preference optimization (DPO)~\cite{rafailov2023dpo,lai2024step_dpo} to equip LLMs with relevance modeling capabilities. However, most existing LLM-based approaches still follow a discriminative optimization paradigm, or else distill LLM knowledge into smaller BERT-style models for deployment. As a result, significant challenges remain in handling difficult search queries — especially those involving deep semantic understanding or multi-step reasoning — in real-world online search engines.


\subsection{Reasoning-Enhanced Post Training}
Reasoning-enhanced post training aims to strengthen models, particularly LLMs, in logical inference and multi-step reasoning, improving performance on tasks involving complex logic, knowledge integration, or long-horizon reasoning. Early approaches relied on supervised learning~\cite{hu2022lora}, training on inputs and ground-truth answers. While effective on small reasoning datasets, these methods generalized poorly and struggled in open- and cross-domain scenarios. The emergence of large-scale pre-trained language models enabled chain-of-thought (CoT)~\cite{wei2022cot}, which guides models to explicitly generate intermediate reasoning steps and decompose complex problems. Subsequently, Tree-of-Thought (ToT)~\cite{yao2023tot} and Bootstrapped Thought~\cite{zelikman2022bot} approaches emerged to further enhance reasoning steps. 

Nevertheless, these methods still lack strong alignment with human preferences. To address this, Reinforcement Learning approaches such as Reinforcement Learning from Human Feedback (RLHF)~\cite{christiano2017RLHF} have been introduced. RLHF typically optimizes LLMs with reward signals that reflect not only answer correctness but also the rationality and interpretability of reasoning steps, reducing logical leaps and hallucinations. Policy optimization algorithms such as Proximal Policy Optimization (PPO)~\cite{schulman2017ppo} are widely adopted for stability and efficiency, while variants like Grouped Rollout Policy Optimization (GRPO)~\cite{shao2024grpo} improve sample efficiency via grouped rollouts and variance reduction. On the other hand, methods such as DPO~\cite{rafailov2023dpo} and Rejection Sampling Fine-tuning~\cite{liu2023rej_sampl_po} have been proposed to bypass explicit reward modeling, directly optimizing from preference data and achieving greater stability and data efficiency. In parallel, process supervision techniques~\cite{lai2024step_dpo} have emerged, optimizing not only final answers but also intermediate reasoning paths, further enhancing generalization across tasks.

\begin{figure*}
    \centering
    \includegraphics[width=0.75\linewidth]{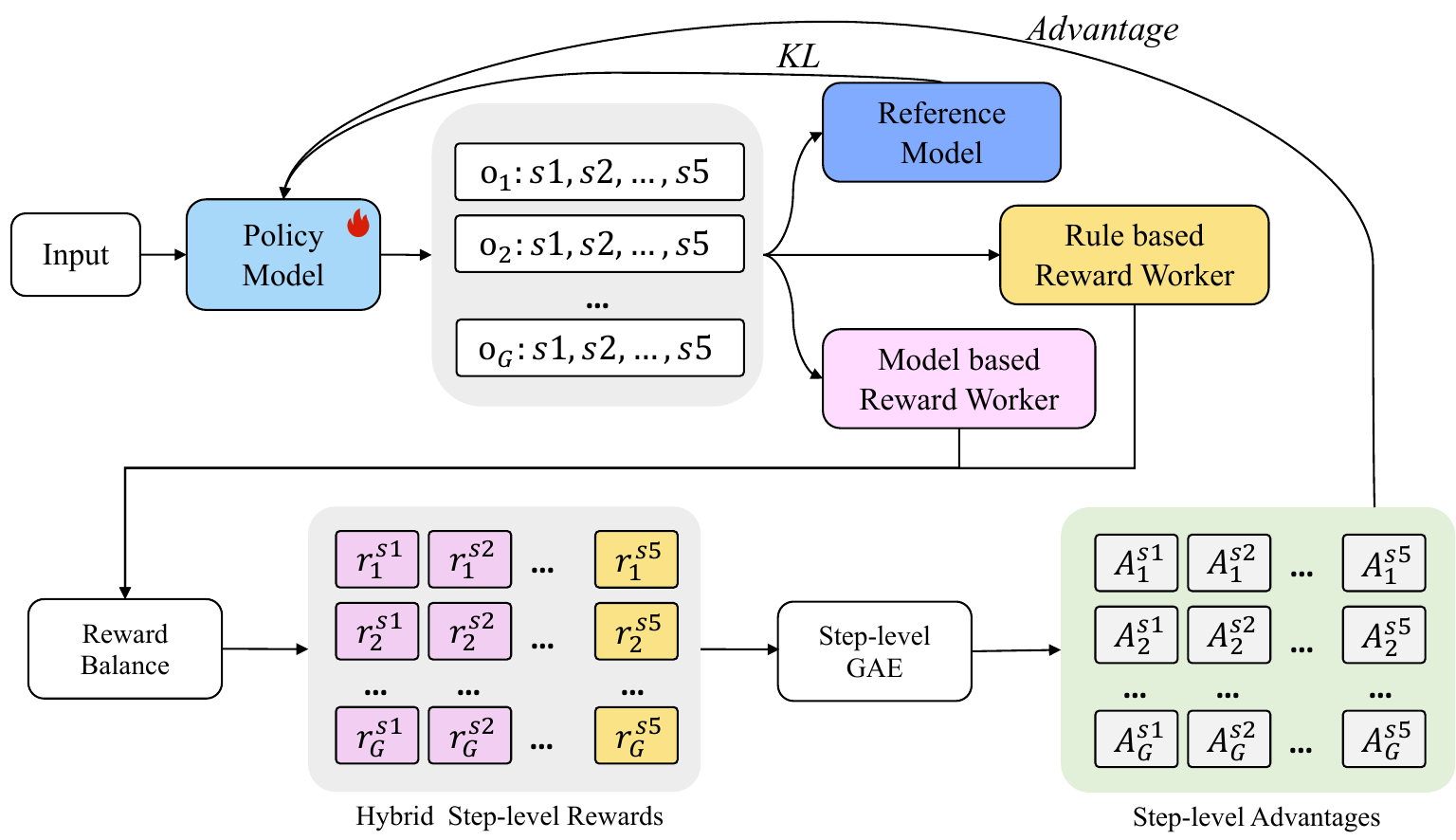}
    \caption{Illustration of our proposed Hybrid Stepwise RL pipeline. Each key step is extracted from the policy-model rollout, and both a generative stepwise reward model and offline human verification are employed to obtain step-level rewards. These rewards are then used to estimate step-level advantages to guide reinforcement learning.}
    \label{fig:srpo}
\end{figure*}

\section{Preliminary}

\subsection{LLM-based E-commerce Relevance Analysis}

To address the "black box" nature of traditional discriminative models in e-commerce search relevance, we reframe the task as an interpretable generative process. Our model, denoted as $\policy$, is trained to generate not only a final relevance label $c$ but also an explicit chain of reasoning $y$. This approach deconstructs the relevance judgment into verifiable intermediate steps, making the model's decision-making process transparent and auditable.

\subsection{Reinforcement Learning with Verifiable Rewards (RLVR)}

\emph{Reinforcement Learning with Verifiable Rewards} (RLVR) refers to a reinforcement learning paradigm in which the reward signal is derived from outcomes that can be independently validated by an external system or mechanism. Normally a verification function $\verification$ that assesses whether a given reward is consistent with the task's specifications.

Formally, for a history or trajectory prefix $\history$, the verification function $\verification: \history \times \reals \rightarrow \{0,1\}$ determines if a reward is valid. The learning objective is therefore to maximize the expected sum of \emph{verified} rewards:
\begin{equation}
    \max_{\pi} \; \expect_{\pi} \left[ \sum_{t=0}^{\infty} \gamma^t r_t \cdot \verification(h_t, r_t) \right],
\end{equation}
where only rewards that pass verification contribute to the agent's return.

\subsubsection{GRPO}

Group Relative Policy Optimization (GRPO)~\cite{shao2024grpo} is one of the RLVR algorithms. Similar to PPO~\cite{schulman2017ppo}, GRPO updates the policy using a clipped surrogate objective. Its key innovation is the use of \emph{grouped rollouts}: for a given input, the policy samples $G$ reasoning trajectories and computes a \emph{relative advantage} for each one based on normalized rewards within the group. This allows the policy to more effectively learn from correct reasoning paths.

The GRPO loss objective is:
\begin{align} \label{eq:grpo}
\mathcal{L}_{\mathrm{GRPO}}(\theta) 
&= -\frac{1}{G} \sum_{i=1}^{G} \frac{1}{|o_i|} \sum_{t=1}^{|o_i|} \bigg[
\min \Big(
\rho_{i, t}(\theta) \, \hat{A}_{i}, \nonumber \\
& \quad \mathrm{clip}\big(
\rho_{i, t}(\theta), \;
1 - \epsilon, \; 1 + \epsilon
\big) \hat{A}_{i}
\Big) - \beta \, \kl\!\left[ \policy \,\|\, \refpolicy \right]
\bigg]
\end{align}
where $\rho_{i, t}(\theta)$ is the importance sampling ratio between the current and old policies, as defined in PPO~\cite{schulman2017ppo}:
\begin{equation}
\rho_{i, t}(\theta) =
\frac{\policy(o_{i,t} \mid q_i, o_{i,<t})}{\oldpolicy(o_{i,t} \mid q_i, o_{i,<t})}
\end{equation}
In GRPO, $\hat{A}_{i}$ is a static, sequence-level advantage, normalized across the group of $G$ rollouts:
\begin{equation} \label{eq:advantage}
\hat{A}_{i} =
\frac{R_i - \mathrm{mean}(\{ R_j \}_{j=1}^G)}{\mathrm{std}(\{ R_j \}_{j=1}^G) + \delta}
\end{equation}
where $R_i$ is the total reward for trajectory $o_i$. This normalization centers and scales the advantage values, promoting relative preference learning and stabilizing the training process.

\section{SHE: A Stepwise Hybrid Examination RL Framework}
In this section, we introduce the Stepwise Hybrid Examination RL Framework (SHE). We detail its core components: data selection strategies (Section~\ref{sec:data_selection}), the training paradigm (Section~\ref{sec:training_paradigm}), and the model architecture and reward design (Section~\ref{sec:model_reward}).

\subsection{Data Selection} \label{sec:data_selection}
\subsubsection{Offline Rejection Sampling}
Rollout-based methods like GRPO rely on reward variance within a group of trajectories to generate a meaningful training signal. If all trajectories for a given input are uniformly correct or incorrect, the relative advantage collapses to zero. To enhance training efficiency, we employ an \textbf{offline rejection sampling} strategy. This involves filtering the training set to discard samples for which the initial policy consistently generates either entirely correct or entirely incorrect reasoning paths, thereby concentrating resources on the most informative samples.

\subsubsection{Data Diversity}
In RL, a strong correlation between high-probability actions and high rewards can lead to policy degeneration, where the output distribution collapses to a few patterns~\cite{yu2025dapo}. To mitigate this, we construct a highly diverse training dataset spanning multiple dimensions, such as industry domains, query types, and relevance grades. This encourages the model to explore a wider range of reasoning strategies and helps prevent policy collapse, promoting generalization.

\subsection{Training Paradigm} \label{sec:training_paradigm}

\subsubsection{Dynamic Difficulty Sampling}
Building upon offline rejection sampling, we introduce a \textbf{dynamic difficulty sampling} mechanism. This method dynamically adjusts the sample selection criteria throughout training. As the policy improves, the definition of a "challenging" sample evolves, enabling the training curriculum to automatically concentrate on the most salient data for the policy's current developmental stage.

\subsubsection{Curriculum Learning}
Our empirical results show that a progressive, multi-stage RL process is more effective than a monolithic approach. We adopt a \textbf{curriculum learning} strategy, where task complexity is gradually increased. This helps mitigate conflicts between different optimization objectives. For example, training may begin with a dataset balanced by query intent, followed by a fine-tuning phase on a dataset balanced by difficulty. Detailed results are presented in Section~\ref{sec:exp_curriculum}.

\subsection{Model and Reward Design} \label{sec:model_reward}
Conventional approaches to e-commerce search relevance often frame the task as a discriminative problem. Such models typically function as black boxes, limiting their interpretability. To enhance transparency and guide the internal reasoning process, we adopt a Chain-of-Thought (CoT) methodology. Given a query and an item, the model generates a step-by-step analysis, culminating in a final relevance grade. We formulate the relevance assessment task as a sequential reasoning process composed of five distinct steps:
\begin{enumerate}[label=(\roman*)]
    \item Query Interpretation ($S_1$),
    \item Item Interpretation ($S_2$),
    \item Category Relevance Evaluation ($S_3$),
    \item Attribute Relevance Evaluation ($S_4$), and
    \item Final Ranking Determination ($S_5$).
\end{enumerate}
Experimental evidence indicates that, when coupled with appropriate techniques, the CoT methodology not only enhances interpretability but can also improve overall performance.

\subsubsection{Generative Stepwise Reward Model Training}

Our multi-step reasoning process includes stages for which ground-truth labels are not readily available (e.g., query interpretation and item interpretation). To provide a reward signal for these indeterminate steps with open-ended responses, we train a generative stepwise reward model, denoted as $\rewardmodel$.

As illustrated in Figure~\ref{fig:main}(A), the training of our reward model proceeds in two stages. First, we perform Supervised Fine-Tuning (SFT). A large corpus of Chains-of-Thought (CoTs) are generated by the policy model, and human annotators provide quality judgments for each reasoning step. This yields a dataset of (query, item, CoT, judgment) tuples, which is used to establish a strong baseline for reward modeling. To further enhance its generalization and ability to handle difficult cases, we then refine the reward model using GRPO. For this stage, we employ difficulty-aware and diversity-aware sampling to generate a subset of challenging examples. Crucially, to ensure high-fidelity labels, these machine-generated samples then undergo meticulous human verification. The resulting high-quality, curated dataset of hard examples is subsequently used to continue training the reward model with GRPO. 

\vspace{0.1in}
\noindent\textbf{Remark 1.} In our experiments, rewards are generated only for the first two reasoning steps, whereas Steps~3 and~4 use human‑annotated labels to achieve higher accuracy. However, obtaining large quantities of such step‑level annotations in every iteration is costly. In these cases, a reward‑only approach serves as a practical alternative and can still deliver competitive performance (see Section~\ref{sec:ablation_hybrid_step} for details).

\vspace{0.1in}
\noindent\textbf{Remark2.} Leveraging the discriminative ability of the reward model, we use it to select valuable examples, as detailed in Section~\ref{sec:rm_for_sampling}.

\begin{figure}
    \centering
    \includegraphics[width=1\linewidth]{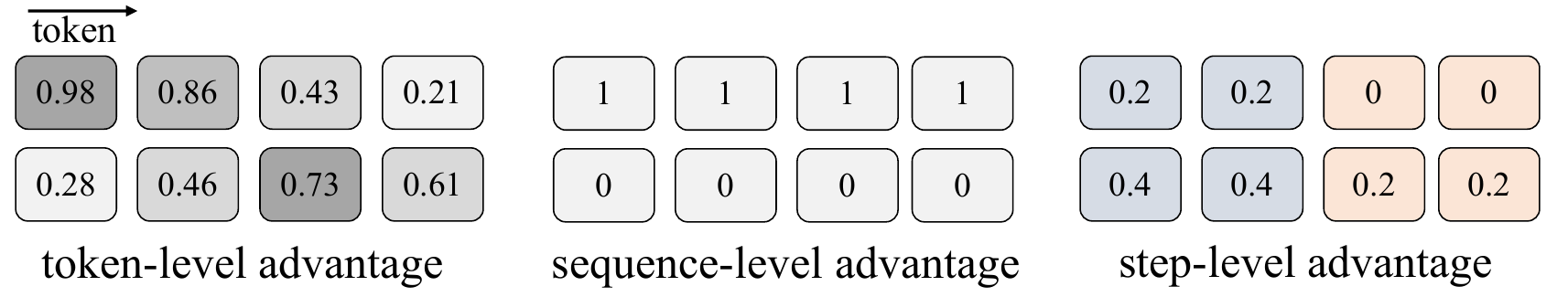}
    \caption{Unlike PPO, which uses token-level advantages, and GRPO, which uses sequence-level advantages, our SRPO estimates step-level advantages.}
    \label{fig:advantage_est}
\end{figure}

\subsubsection{Stepwise Hybrid Reward}
Each step $S_j$ in the reasoning chain produces an intermediate conclusion that can be individually verified, allowing for a step-specific reward signal. The verifiability of these steps is heterogeneous. For structured tasks like category ($S_3$) and attribute ($S_4$) matching, reliable ground truth can be pre-computed. In contrast, for open-ended semantic tasks like query ($S_1$) and item ($S_2$) interpretation, defining correctness \emph{a priori} is infeasible.

To address this, we introduce a \textbf{Stepwise Hybrid Reward} scheme. Let $o_i \in \outputspace$ be the $i$-th model-generated sequence for a given query $q \in \queryspace$ and item $p \in \itemspace$. For any step $S_j$, let $\pi_{S_j}: \outputspace \rightarrow \stepoutputspace_j$ be a deterministic \emph{parsing operator} that extracts the predicted output $\hat{y}_j = \pi_{S_j}(o_i)$ for that step from the full sequence. The stepwise rewards $\{r_i^{S_j}\}$ are defined as:
\begin{equation}
\label{eq:stepwise-reward}
r_{i}^{S_j} = 
\begin{cases} 
    \rewardmodel(q, p, \pi_{S_j}(o_i)) & \text{if } j \in \{1,2\} \\
    \indicator\!\left[ \pi_{S_j}(o_i) = gt_j \right] & \text{if } j \in \{3,4\}
\end{cases}
\end{equation}
where $\indicator[\cdot]$ is the indicator function. The reward for the final step, $S_5$, can be independently defined based on the correctness of the final relevance grade. 
The total reward for a trajectory is the sum of its step rewards: $R_i = \sum_{j=1}^{J} r_i^{S_j}$.

\begin{algorithm}[t]
    \caption{SRPO: RL with Stepwise Hybrid Reward}
    \label{alg:srpo}          
    \begin{algorithmic}[1] 
        \Require Policy model $\policy$, reward model $\rewardmodel$
        \Require Ground truth $\{gt_3, gt_4\}$, dataset $\dataset$ 
        \Require Discount factor $\gamma$
        \Ensure Optimized policy parameters $\theta$

        \While{not converged}
            \State Select minibatch $B \subset \dataset$ via dynamic difficulty sampling
            \ForAll{$(q, p) \in B$}
                \State Generate $G$ rollouts $\{o_i\}_{i=1}^G$ from $\oldpolicy$
                \ForAll{rollout $o_i$ in $\{o_i\}_{i=1}^G$}
                    \For{$j = 1$ to $J$} \Comment{Calculate step rewards}
                        \State $\hat{y}_{i,j} \gets \pi_{S_j}(o_i)$ 
                        \State $r_i^{S_j} \gets \text{GetReward}(S_j, \hat{y}_{i,j}, q, p, \rewardmodel, gt_j)$ (Eq.~\ref{eq:stepwise-reward})
                    \EndFor
                    \For{$t = 1$ to $|o_i|$} \Comment{Compute step‑level advantage and assign to tokens in this step}
                        \State Let $S_j$ be the step corresponding to token $o_{i,t}$
                        \State $A_i(t) \gets \sum_{k=j}^{J} (\gamma)^{k-j} \, r_i^{S_k}$
                    \EndFor
                \EndFor
            \EndFor
            \State Update $\policy$ using the SRPO loss from Eq.~\ref{eq:srpo_loss} with $\{A_i(t)\}$
        \EndWhile

        \State \Return $\policy$
    \end{algorithmic}
\end{algorithm}

\subsubsection{SRPO: Stepwise Reward Policy Optimization}
In PPO~\cite{schulman2017ppo}, token-level advantages are typically used; however, obtaining token-level advantages requires estimating $V(s_t)$, which can be challenging to train and computationally expensive. By contrast, GRPO~\cite{shao2024grpo} assigns a single scalar reward at the sequence level and propagates it uniformly to all tokens. This is suboptimal for multi-step reasoning: correct intermediate steps may be unfairly penalized if the final outcome is incorrect, and erroneous steps may be reinforced if the final outcome is correct by chance. Consequently, credit assignment becomes noisy, and reward hacking becomes more likely.

To mitigate this, we introduce Stepwise Reward Policy Optimization (SRPO), which enhances the GRPO framework by integrating step-level rewards directly into the advantage estimation. See Figure~\ref{fig:advantage_est} for the contrast between PPO, GRPO, and SRPO, and Figure~\ref{fig:srpo} for the overall SRPO pipeline. Unlike token-level or sequence-level advantages, SRPO computes a step-level advantage $A_i(t)$ for token $i$ at position $t$ that belongs to reasoning step $S_j$. This advantage is the discounted sum of rewards from the current and subsequent steps:
$$
A_i(t) = \sum_{k=j}^{J} (\gamma)^{k-j} \, r_i^{S_k},
$$
where $J$ is the total number of reasoning steps, $t$ is the token index, $i$ is the rollout sequence index, and $\gamma$ is the discount factor. The term $r_i^{S_k}$ denotes the reward for step $S_k$ of sequence $i$. For a token in step $S_j$ in sequence $i$, this formulation captures its cumulative stepwise advantage over all remaining reasoning steps.


This refined, time-dependent advantage $A_i(t)$ replaces the static advantage $\hat{A}_i$ in the GRPO objective. The final SRPO loss is:
\begin{align} \label{eq:srpo_loss}
\mathcal{L}_{\mathrm{SRPO}}(\theta) 
&= -\frac{1}{G} \sum_{i=1}^{G} \frac{1}{|o_i|} \sum_{t=1}^{|o_i|} \bigg[
 \min \Big(
\rho_{i, t}(\theta) \, A_i(t), \\
& \quad \mathrm{clip}\big(
\rho_{i, t}(\theta), \;
1 - \epsilon, \; 1 + \epsilon
\big) A_i(t)
\Big) - \beta \, \kl\!\left[ \policy \,\|\, \refpolicy \right]
\bigg] \nonumber
\end{align} 
This approach enables more targeted credit assignment and efficient error correction throughout the reasoning chain.

\begin{table*}[!t]
\setlength{\tabcolsep}{15pt}
\caption{Offline evaluation results for baseline methods and the proposed approach, reported on the in-the-wild test set.}
\label{tbl:main}
\centering
\begin{tabular}{lrrrrr}
\toprule
\textbf{Method} & \textbf{Class-1 F1} & \textbf{Class-2 F1} & \textbf{Class-3 F1} & \textbf{Macro F1} & \textbf{Accuracy} \\
\midrule
SFT & 43.15 & \textbf{64.23} & 84.31 & 63.9  & 76.41 \\
DPO & 44.59 & 63.12 & 86.32 & 64.67 &  78.32 \\
GRPO & 45.41 & 62.95 & 86.49 & 64.95 &  78.47 \\
\midrule
SRPO & \textbf{47.44} & 63.78 & \textbf{86.87} & \textbf{66.03}  & \textbf{79.18} \\
\bottomrule
\end{tabular}
\end{table*}



\section{Experiments}

\subsection{Experimental Setup}

\subsubsection{Dataset} \label{sec:dataset}
Our training set is derived from Taobao’s online search logs, consisting of sampled candidate query–item pairs evaluated along two dimensions: 1) difficulty, and 2) data diversity. 

Our \textbf{in-the-wild test set} is derived from Taobao’s online search logs, sampled from the candidate space scored by the relevance model. It contains 21,616 manually annotated query–item pairs. To rigorously assess the model’s reasoning ability, the query distribution is deliberately focused on four challenging categories: \textit{question answering (Q\&A)}, \textit{alternatives}, \textit{negation}, and \textit{knowledge-based} queries. Representative examples are shown in Table~\ref{tbl:online_results}. Additionally, we utilize a \textbf{balanced query-type test set} comprising 77{,}380 samples to support our Diverse Sampling and Curriculum Learning ablation studies.

\subsubsection{Evaluation Protocol} \label{sec:metric}
\textbf{Offline evaluation: }
We categorize the relevance between a user query and a product item into three distinct levels: \textit{Bad} (Class 1), \textit{Mid} (Class 2), and \textit{Good} (Class 3)\footnote{The "Good" category comprises two sub-classes in the raw data, we consolidate them into a single class for online deployment.}. To provide a comprehensive assessment, model performance is reported using class-specific \textbf{F1 scores}. Additionally, we compute the \textbf{Macro-F1} and \textbf{Overall Accuracy} across all three classes to evaluate the global classification effectiveness. 



\textbf{Online evaluation:} We conduct live A/B experiments with human judgments to compute the following metrics:

\begin{itemize}[leftmargin=10pt]
  \item \textbf{GSB (Good/Same/Bad):} A side-by-side comparison metric used in A/B testing. Human assessors evaluate outputs from two competing systems for the same query. \(\mathrm{GSB} + x\%\) indicates that, in \(x\%\) of comparisons, the test system is preferred over the baseline.
  \item \textbf{Query Goodrate:} A page-level relevance metric measured as the proportion of queries whose returned pages are rated \emph{Good} or \emph{Mid} on the four-point scale. Ratings are based on page-level relevance across the displayed items.
  \item \textbf{Item Goodrate:} An item-level relevance metric defined as the mean proportion of items per request rated as \textit{Class 3} or \textit{Class 4}. The mean item goodrate is reported across all evaluation requests.
\end{itemize}

\subsubsection{Base Model}
We use \texttt{Tbstar-42B-A3.5}, a proprietary Mixture-of-Experts (MoE) model developed internally at Taobao for all baselines and our method. It was pre-trained from scratch on over 10 trillion tokens of general and e-commerce-specific data, comprising 42B total parameters, with 3.5B active during inference.

\subsubsection{Baselines} \label{sec:baselines}
We benchmark our approach against the following strong baselines on \texttt{Tbstar-42B-A3.5} models:
\begin{itemize}[leftmargin=10pt]
    \item \textbf{\texttt{Tbstar-SFT}}: The \texttt{Tbstar-SFT} baseline is obtained by fine-tuning this model on a dataset of 2 million samples using Supervised Fine-Tuning (SFT).
    \item \textbf{\texttt{Tbstar-DPO}}: This model is initialized from the \texttt{Tbstar-SFT} checkpoint and subsequently fine-tuned using the DPO algorithm on a 300k-sample dataset of pairwise preference labels.
    \item \textbf{\texttt{Tbstar-GRPO}}: To create a strong reinforcement learning baseline, this model is initialized from the \texttt{Tbstar-DPO} checkpoint and further fine-tuned using the GRPO algorithm with 39,161 samples.
    \item \textbf{\texttt{Tbstar-GRPO*}}: Extends \texttt{Tbstar-GRPO} by incorporating stepwise verification rewards for Steps~3 and~4 into the final reward (see Section~\ref{sec:model_reward} for details), while keeping all other settings identical to \texttt{Tbstar-DPO}.  
\end{itemize}

\subsubsection{Reinforcement-Learning Configuration}
For RL training (including GRPO and our SRPO), the rollout batch size is 64, batch normalization is applied, and the learning rate is set to \(1.0\times 10^{-6}\). For each query, we generate 16 responses using sampling with a temperature \(\tau = 0.99\) and top-\(k = 100\). Difficulty-sampling thresholds are set from 0.01 (minimum) to 0.90 (maximum). For loss-target clipping, the clipping magnitude is 0.2, i.e., values are restricted to \([-0.2, 0.2]\). Advantage values are clipped at a magnitude of 2. All experiments are implemented using the open-source RL framework ROLL \cite{wang2025roll}.

\begin{table}[]
\caption{The performance of Generative Stepwise Reward Model.}
\label{tbl:reward_model_performance}
\begin{tabular}{lllll}
\toprule
     & step1 & step2 & step3 & step4 \\
\midrule
+SFT        & 86.92      &86.21 &89.91 &83.60  \\
+RL (GRPO)  &   89.84    &  89.10     &  90.22     & 86.36     \\
\bottomrule
\end{tabular}
\end{table}

\begin{table*}[]
\caption{Ablation study on the components of our stepwise reward mechanism. We compare a standard sequence-level reward baseline against our step-level rewards using only a reward model and our proposed hybrid approach. SRPO consistently improves performance in both settings. Results are reported on the in-the-wild dataset.}
\label{tbl:ablation_step_level}
\begin{tabular}{lcccccc}
\toprule
\textbf{Method}        & \textbf{Class-1 F1} & \textbf{Class-2 F1} & \textbf{Class-3 F1}  & \textbf{Macro F1} & \textbf{Accuracy} \\
\midrule
GRPO & 45.41 & 62.95 & 86.49 & 64.95 &  78.47 \\
\midrule
\textit{Reward Model Only Setting} \\
\; + Step Reward (Reward Model Only) &   45.73  &  63.82  & 86.63         &   65.39       &   78.77       \\
\; + Step GAE    & 46.6 & 63.98 & 86.68         &  65.76        &  78.88    \\
\midrule
\textit{Hybrid Reward Setting} \\
\; + Step Reward (Hybrid) & 46.4 & \textbf{64.58} & 86.47        & 65.82         &    78.71      \\
\; + Step GAE   & \textbf{47.44} & 63.78 & \textbf{86.87} & \textbf{66.03}  & \textbf{79.18} \\
\bottomrule
\end{tabular}
\end{table*}



\subsubsection{Generative Stepwise Reward Model Training Details}
We employ a 13B dense model, Tbstar-13B, as our reward model. We adopt a two-stage training scheme: (i) supervised fine-tuning (SFT) and (ii) reinforcement learning with GRPO. For SFT of the Generative Stepwise Reward Model, we use a collection of CoT-Judge pairs, totaling 30{,}000 samples. In addition, we sample 13{,}000 instances to support RL-based training of the reward model using GRPO.

\subsubsection{Policy Model Training Details}  
To evaluate the proposed approach, we use the \texttt{Tbstar-42B-A3.5} MoE model as the policy model. Training begins from the \texttt{Tbstar-DPO} checkpoint (see Section~\ref{sec:baselines} for details) and leverages the 39{,}161-sample dataset described in Section~\ref{sec:dataset}. In all experiments, the discount factor of our proposed approach is set to $\gamma = 1$.  
Detailed analysis regarding the computational overhead of SRPO is provided in Appendix~\ref{app:computaion_overhead}.


\subsection{Main Results}
Table\ref{tbl:main} reports the offline evaluation results. All RL-based methods substantially improve the SFT baseline in terms of overall accuracy and Macro F1. Compared with the baselines, SRPO outperforms SFT, DPO, and GRPO in Class-1 F1, Class-3 F1, Macro F1, and Accuracy. Concretely, SHE achieves Class-1 F1 = 47.44, Class-2 F1 = 63.78, Class-3 F1 = 86.87, Macro F1 = 66.03, and Accuracy = 79.18; by comparison, the best baseline (GRPO) attains Class-1 F1 = 45.41, Class-2 F1 = 62.95, Class-3 F1 = 86.49, Macro F1 = 64.95, and Accuracy = 78.47. It is worth noting that while SRPO does not have the top Class-2 F1 (64.23 achieved by SFT), it achieves superior performance across all other main evaluation metrics, validating the effectiveness of the proposed approach in offline assessments.

\begin{table}[t]
\centering
\caption{Ablation study on diverse sampling strategies, evaluated on the balanced query-type test set.}
\label{tbl:ablation_data_diversity}
\begin{tabular}{lc}
\toprule
\textbf{Method} & \textbf{Macro F1} \\
\midrule
Baseline & 65.02 \\
GRPO w/o diverse sampling & 65.64 \\
GRPO w diverse sampling & \textbf{66.49} \\
\bottomrule
\end{tabular}
\end{table}

\subsection{Performance of Reward Model}
The performance evaluation of the reward model, as detailed in Table~\ref{tbl:reward_model_performance}, showcases its high degree of accuracy and reliability in verifying individual reasoning steps. The model consistently achieves strong performance across a four-step process, with verification accuracies reaching as high as 90.22\%. Even at its lowest point, the model maintains a robust accuracy of 86.36\%, demonstrating its consistent ability to accurately assess the correctness of stepwise outputs. 

\subsection{Ablation Study}
In this section, we conduct ablation studies to systematically evaluate the contributions of the key components of our proposed framework. For the ablation of Diverse Sampling and Curriculum Learning, we report the results on the balanced query-type test set (see Section~\ref{sec:dataset}) for details.

\begin{table}[t]
\centering
\caption{Multi-stage curriculum learning strategy, evaluated on the balanced query-type test set.}
\label{tbl:ablation_curriculum_learning}
\begin{tabular}{p{0.36\textwidth} c}
\toprule
\textbf{Training Stage} & \textbf{Macro F1} \\
\midrule
Baseline & 65.02 \\
Stage 1a: GRPO w easier label-balance data & 66.49 \\
Stage 1b: GRPO w harder label-balance data & 66.81 \\
Stage 1c: GRPO w full label-balance data & 66.52 \\
\midrule
Stage 2: GRPO w easier-then-harder curriculum & \textbf{67.12} \\
\bottomrule
\end{tabular}
\end{table}

\begin{table}[t]
\centering
\caption{Multi-stage curriculum across additional data types, evaluated on the balanced query-type test set.}
\label{tbl:ablation_curriculum_learning2}
\begin{tabular}{p{0.36\textwidth} c}
\toprule
\textbf{Training Stage} & \textbf{Macro F1}  \\
\midrule
Baseline & 60.01  \\
Stage 1a: GRPO w Diverse Query-type data & 63.20  \\
Stage 1b: GRPO w Diverse label data & 65.84  \\
Stage 1c: GRPO w full data & 63.94  \\
\midrule
Stage 2: Diverse Query-type then diverse label & \textbf{66.84} \\
\bottomrule
\end{tabular}
\end{table}

\begin{table*}[h]
\centering
\caption{Side-by-Side Human Evaluations}
\label{tbl:online_results}
\setlength{\tabcolsep}{8pt}
\begin{tabular}{p{2cm}p{6cm}ccc}
\toprule
\textbf{Query Type} & \textbf{Case} & \textbf{GSB} & \textbf{Query Goodrate} & \textbf{Item Goodrate} \\
\midrule
Q\&A & What to wear for mountaineering? & +12.91\% & +3.33pt & +3.89pt \\
Alternative & Arc'teryx alternative & +0.61\% & +0.72pt & +2.83pt \\
Negative & Non-turtleneck sweater & +5.85\% & +1.29pt & +2.42pt \\
Knowledge & Grease-removing products for the kitchen & +3.47\% & +0.93pt & +1.02pt \\
\bottomrule
\end{tabular}
\end{table*}

\begin{table}[h]
\centering
\caption{
Performance and data efficiency of Reward Model (RM)-based data selection, evaluated on the balanced query-type test set.}
\label{tbl:rm_for_data_selection}
\begin{tabular}{lcc}
\toprule
\textbf{Method} & \textbf{Training Samples} & \textbf{Macro F1} \\
\midrule
GRPO w Full data & 4w & 65.64 \\
GRPO w RM-selected subset & 2w & 65.54 \\
\bottomrule
\end{tabular}
\end{table}

\subsubsection{Hybrid Stepwise Reward} \label{sec:ablation_hybrid_step}
We first evaluate the effectiveness of our stepwise reward design by comparing it against a standard GRPO baseline that uses only a sequence-level reward. As shown in Table~\ref{tbl:ablation_step_level}, our approach is analyzed in two main settings: one using only the reward model for all steps ("Reward Model Only") and our proposed hybrid approach ("Hybrid").

The results clearly demonstrate that providing step-level rewards consistently outperforms the baseline, confirming the benefits of fine-grained credit assignment. Furthermore, integrating SRPO yields additional performance gains in both settings, highlighting its effectiveness in propagating rewards accurately through the reasoning chain. Notably, our full hybrid model combined with SRPO achieves the best overall performance.

\subsubsection{Diverse Sampling}
We study how different data diversity strategies affect model performance. As shown in Table~\ref{tbl:ablation_data_diversity}, we compare a no-diversity baseline with strategies that promote diversity across query intents, domains, and class labels. The results show that each diversity strategy improves Macro F1, with label-based diverse sampling providing the largest gains. 


\subsubsection{Curriculum Learning} \label{sec:exp_curriculum}  
We evaluate the effectiveness of our multi-stage curriculum learning strategy.
First, we construct an easier-then-harder curriculum: the easier dataset contains 30k samples and the harder dataset contains 1.3k samples, selected through difficulty-based and diversity-based sampling. As shown in Table~\ref{tbl:ablation_curriculum_learning}, this two-stage curriculum yields superior performance compared to: stage~1a (training only on the easier, label-balanced data), stage~1b (GRPO training on the harder, label-balanced data), and stage~1c (training on the full dataset).
Next, we examine a multi-stage curriculum using datasets with different balancing strategies. The results are shown in Table~\ref{tbl:ablation_curriculum_learning2}.  
Stage~1a, Stage~1b and Stage~1c correspond to single training stages, while Stage~2 adopts a two-stage curriculum: first balancing by query types, then by labels. This two-stage approach outperforms both single-stage settings.

\subsection{Reward Model for Efficient Sample Selection} \label{sec:rm_for_sampling}
To improve training efficiency, we deploy a pre-trained reward model to guide data selection, prioritizing high-value samples. The reward model estimates sample quality and can flag samples considered suboptimal, capturing both process-level and outcome-level errors. Selected samples are then used for reinforcement learning.

As shown in Table~\ref{tbl:rm_for_data_selection}, this approach reduces the required training data while preserving performance. Specifically, a curated subset of approximately 20k samples yields a Macro F1 of 65.54, closely matching the baseline with about 40k samples (65.64). This demonstrates data-efficient learning enabled by reward-based sampling.

\subsection{Online Evaluation}
\subsubsection{Human Evaluation}
To evaluate the model’s effectiveness in practical settings, we conducted a side-by-side comparison against human annotations,
Table~\ref{tbl:online_results} . For 2,000 queries, results from both experimental systems were obtained, and their top-10 outputs were assessed using three distinct metrics (details in Section~\ref{sec:metric}). 
The relevance improvements reported in Table~\ref{tbl:online_results}. 

\subsubsection{Online Deployment} Only the policy model is deployed for online inference. 
To ensure low-latency and stable performance, we follow the label-pioneering method~\cite{dong2025taosr1} by decoding only the first token. Combined with model quantization, this optimization reduces p99 latency to below 400 ms.

\subsubsection{Online Performance}
Despite human evaluations indicating improved relevance quality, initial online deployment revealed a modest decline in core business indicators—specifically, directly attributed order count and gross merchandise value (GMV)—relative to the baseline. A root-cause analysis determined that the upstream recall stage predominantly retrieved items that were relevant but historically associated with low sales. While such items were correctly judged as relevant by SHE, they exhibited limited conversion potential. 
To address this mismatch, we enhanced the upstream pipeline by (i) introducing multi-path recall incorporating personalized signals, and (ii) adding a pre-ranking stage jointly optimized for both relevance and conversion likelihood. These modifications substantially increased the sales efficiency of the candidate set prior to fine-grained relevance scoring.
Following these optimizations, business metrics were restored to levels statistically indistinguishable from the baseline. A four‑day A/B test(4\% traffic) showed gains in Direct Clean GMV(+1.48\%), Orders(+1.26\%), IPV(+1.15\%), and Exposed PV(+2.37\%), while maintaining efficiency.

\section{Conclusion}
While reasoning LLMs offer transformative potential for e-commerce search relevance, they pose critical RLVR challenges: reward sparsity and equitable credit assignment.
To address these, we propose SHE, a reinforcement learning framework that integrates generative stepwise rewards with offline human verification. By optimizing across distinct reasoning stages, SHE enhances efficiency while minimizing computational costs. Central to our framework is Stepwise Reward Policy Optimization (SRPO), which employs step-level supervision to mitigate reward sparsity in complex reasoning tasks. Furthermore, by incorporating Difficulty-aware Sampling, Diverse Sampling, and Curriculum Learning, our approach substantially improves RL convergence and performance. Extensive experiments in real-world e-commerce scenarios demonstrate the superior effectiveness of the proposed framework.



\bibliographystyle{ACM-Reference-Format}  
\bibliography{main}

@String{Chelsea = "Chelsea" }

@article{christiano2017RLHF,
  title={Deep reinforcement learning from human preferences},
  author={Christiano, Paul F and Leike, Jan and Brown, Tom and Martic, Miljan and Legg, Shane and Amodei, Dario},
  journal={Advances in neural information processing systems},
  volume={30},
  year={2017}
}

@article{singhal2001modern_ir_tfidf,
  title={Modern information retrieval: A brief overview},
  author={Singhal, Amit and others},
  journal={IEEE Data Eng. Bull.},
  volume={24},
  number={4},
  pages={35--43},
  year={2001}
}

@article{hambarde2023IR_advances_and_beyond_repreasentation_based,
  title={Information retrieval: recent advances and beyond},
  author={Hambarde, Kailash A and Proenca, Hugo},
  journal={IEEE Access},
  volume={11},
  pages={76581--76604},
  year={2023},
  publisher={IEEE}
}

@article{robertson2009BM25,
  title={The probabilistic relevance framework: BM25 and beyond},
  author={Robertson, Stephen and Zaragoza, Hugo and others},
  journal={Foundations and Trends{\textregistered} in Information Retrieval},
  volume={3},
  number={4},
  pages={333--389},
  year={2009},
  publisher={Now Publishers, Inc.}
}

@inproceedings{karmaker2017application,
  title={On application of learning to rank for e-commerce search},
  author={Karmaker Santu, Shubhra Kanti and Sondhi, Parikshit and Zhai, ChengXiang},
  booktitle={Proceedings of the 40th international ACM SIGIR conference on research and development in information retrieval},
  pages={475--484},
  year={2017}
}

@article{vaswani2017attention,
  title={Attention is all you need},
  author={Vaswani, Ashish and Shazeer, Noam and Parmar, Niki and Uszkoreit, Jakob and Jones, Llion and Gomez, Aidan N and Kaiser, {\L}ukasz and Polosukhin, Illia},
  journal={Advances in neural information processing systems},
  volume={30},
  year={2017}
}

@inproceedings{devlin2019bert,
  title={Bert: Pre-training of deep bidirectional transformers for language understanding},
  author={Devlin, Jacob and Chang, Ming-Wei and Lee, Kenton and Toutanova, Kristina},
  booktitle={Proceedings of the 2019 conference of the North American chapter of the association for computational linguistics: human language technologies, volume 1 (long and short papers)},
  pages={4171--4186},
  year={2019}
}

@inproceedings{yao2022reprbert,
  title={ReprBERT: distilling BERT to an efficient representation-based relevance model for e-commerce},
  author={Yao, Shaowei and Tan, Jiwei and Chen, Xi and Zhang, Juhao and Zeng, Xiaoyi and Yang, Keping},
  booktitle={Proceedings of the 28th ACM SIGKDD conference on knowledge discovery and data mining},
  pages={4363--4371},
  year={2022}
}

@article{schulman2017ppo,
  title={Proximal policy optimization algorithms},
  author={Schulman, John and Wolski, Filip and Dhariwal, Prafulla and Radford, Alec and Klimov, Oleg},
  journal={arXiv preprint arXiv:1707.06347},
  year={2017}
}

@article{rafailov2023dpo,
  title={Direct preference optimization: Your language model is secretly a reward model},
  author={Rafailov, Rafael and Sharma, Archit and Mitchell, Eric and Manning, Christopher D and Ermon, Stefano and Finn, Chelsea},
  journal={Advances in neural information processing systems},
  volume={36},
  pages={53728--53741},
  year={2023}
}

@article{skalse2022reward_game,
  title={Defining and characterizing reward gaming},
  author={Skalse, Joar and Howe, Nikolaus and Krasheninnikov, Dmitrii and Krueger, David},
  journal={Advances in Neural Information Processing Systems},
  volume={35},
  pages={9460--9471},
  year={2022}
}

@article{hu2022lora,
  title={Lora: Low-rank adaptation of large language models.},
  author={Hu, Edward J and Shen, Yelong and Wallis, Phillip and Allen-Zhu, Zeyuan and Li, Yuanzhi and Wang, Shean and Wang, Lu and Chen, Weizhu and others},
  journal={ICLR},
  volume={1},
  number={2},
  pages={3},
  year={2022}
}

@article{wei2022cot,
  title={Chain-of-thought prompting elicits reasoning in large language models},
  author={Wei, Jason and Wang, Xuezhi and Schuurmans, Dale and Bosma, Maarten and Xia, Fei and Chi, Ed and Le, Quoc V and Zhou, Denny and others},
  journal={Advances in neural information processing systems},
  volume={35},
  pages={24824--24837},
  year={2022}
}

@article{zelikman2022bot,
  title={Star: Bootstrapping reasoning with reasoning},
  author={Zelikman, Eric and Wu, Yuhuai and Mu, Jesse and Goodman, Noah},
  journal={Advances in Neural Information Processing Systems},
  volume={35},
  pages={15476--15488},
  year={2022}
}

@article{liu2023rej_sampl_po,
  title={Statistical rejection sampling improves preference optimization},
  author={Liu, Tianqi and Zhao, Yao and Joshi, Rishabh and Khalman, Misha and Saleh, Mohammad and Liu, Peter J and Liu, Jialu},
  journal={arXiv preprint arXiv:2309.06657},
  year={2023}
}

@misc{yao2023tot,
      title={Tree of Thoughts: Deliberate Problem Solving with Large Language Models}, 
      author={Shunyu Yao and Dian Yu and Jeffrey Zhao and Izhak Shafran and Thomas L. Griffiths and Yuan Cao and Karthik Narasimhan},
      year={2023},
      eprint={2305.10601},
      archivePrefix={arXiv},
      primaryClass={cs.CL},
      url={https://arxiv.org/abs/2305.10601}, 
}

@article{lai2024step_dpo,
  title={Step-dpo: Step-wise preference optimization for long-chain reasoning of llms},
  author={Lai, Xin and Tian, Zhuotao and Chen, Yukang and Yang, Senqiao and Peng, Xiangru and Jia, Jiaya},
  journal={arXiv preprint arXiv:2406.18629},
  year={2024}
}

@article{shao2024grpo,
  title={Deepseekmath: Pushing the limits of mathematical reasoning in open language models},
  author={Shao, Zhihong and Wang, Peiyi and Zhu, Qihao and Xu, Runxin and Song, Junxiao and Bi, Xiao and Zhang, Haowei and Zhang, Mingchuan and Li, YK and Wu, Yang and others},
  journal={arXiv preprint arXiv:2402.03300},
  year={2024}
}

@inproceedings{tang2025lref,
  title={LREF: A Novel LLM-based Relevance Framework for E-commerce Search},
  author={Tang, Tian and Tian, Zhixing and Zhu, Zhenyu and Wang, Chenyang and Hu, Haiqing and Tang, Guoyu and Liu, Lin and Xu, Sulong},
  booktitle={Companion Proceedings of the ACM on Web Conference 2025},
  pages={468--475},
  year={2025}
}

@article{wang2025roll,
  title={Reinforcement Learning Optimization for Large-Scale Learning: An Efficient and User-Friendly Scaling Library},
  author={Wang, Weixun and Xiong, Shaopan and Chen, Gengru and Gao, Wei and Guo, Sheng and He, Yancheng and Huang, Ju and Liu, Jiaheng and Li, Zhendong and Li, Xiaoyang and others},
  journal={arXiv preprint arXiv:2506.06122},
  year={2025}
}

@article{yu2025dapo,
  title={Dapo: An open-source llm reinforcement learning system at scale},
  author={Yu, Qiying and Zhang, Zheng and Zhu, Ruofei and Yuan, Yufeng and Zuo, Xiaochen and Yue, Yu and Dai, Weinan and Fan, Tiantian and Liu, Gaohong and Liu, Lingjun and others},
  journal={arXiv preprint arXiv:2503.14476},
  year={2025}
}

@article{chen2024proRBP,
  title={Towards Boosting LLMs-driven Relevance Modeling with Progressive Retrieved Behavior-augmented Prompting},
  author={Chen, Zeyuan and Wu, Haiyan and Wu, Kaixin and Chen, Wei and Zhong, Mingjie and Xu, Jia and Liu, Zhongyi and Zhang, Wei},
  journal={arXiv preprint arXiv:2408.09439},
  year={2024}
}

@article{achiam2023gpt4,
  title={Gpt-4 technical report},
  author={Achiam, Josh and Adler, Steven and Agarwal, Sandhini and Ahmad, Lama and Akkaya, Ilge and Aleman, Florencia Leoni and Almeida, Diogo and Altenschmidt, Janko and Altman, Sam and Anadkat, Shyamal and others},
  journal={arXiv preprint arXiv:2303.08774},
  year={2023}
}

@article{yang2025qwen3,
  title={Qwen3 technical report},
  author={Yang, An and Li, Anfeng and Yang, Baosong and Zhang, Beichen and Hui, Binyuan and Zheng, Bo and Yu, Bowen and Gao, Chang and Huang, Chengen and Lv, Chenxu and others},
  journal={arXiv preprint arXiv:2505.09388},
  year={2025}
}

@article{mehrdad2024llm4relevance,
  title={Large language models for relevance judgment in product search},
  author={Mehrdad, Navid and Mohapatra, Hrushikesh and Bagdouri, Mossaab and Chandran, Prijith and Magnani, Alessandro and Cai, Xunfan and Puthenputhussery, Ajit and Yadav, Sachin and Lee, Tony and Zhai, ChengXiang and others},
  journal={arXiv preprint arXiv:2406.00247},
  year={2024}
}

@article{dong2025taosr1,
  title={TaoSR1: The Thinking Model for E-commerce Relevance Search},
  author={Dong, Chenhe and Yao, Shaowei and Jiao, Pengkun and Yang, Jianhui and Jin, Yiming and Huang, Zerui and Zhou, Xiaojiang and Ou, Dan and Tang, Haihong},
  journal={arXiv preprint arXiv:2508.12365},
  year={2025}
}

@article{turpin2023llm_don_always,
  title={Language models don't always say what they think: Unfaithful explanations in chain-of-thought prompting},
  author={Turpin, Miles and Michael, Julian and Perez, Ethan and Bowman, Samuel},
  journal={Advances in Neural Information Processing Systems},
  volume={36},
  pages={74952--74965},
  year={2023}
}

\clearpage

\appendix


\begin{figure*}[hb]
\centering
\includegraphics[width=0.9\linewidth]{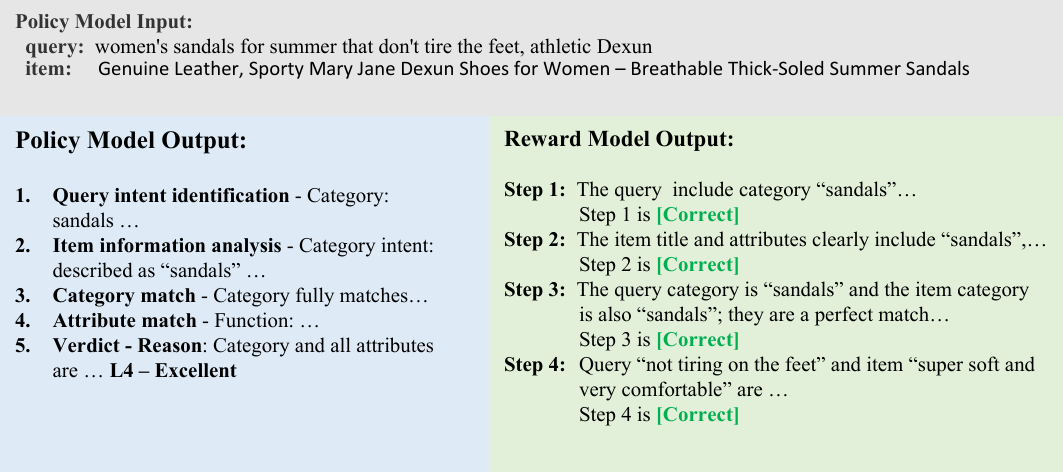}
\caption{Representative outputs produced by our generative stepwise reward model.}
\label{fig:rm_output_examples}
\end{figure*}

\section{Generative Reward Model V.S. Scalar Reward Model}
We compare three reward-modeling approaches within our RLHF pipeline: (i) the vanilla GPRO baseline, (ii) a Scalar Reward Model (SRM), and (iii) a Generative Reward Model (GRM). Evaluations are conducted on 13k-sample data collected from online data. Results are summarized in Table~\ref{tab:reward_results}. The Generative Reward Model achieves the best performance in both settings. In addition, the Scalar Reward Model exhibits interpretability challenges: we cannot identify the reasons behind the scores, which limits further analysis and diagnostic refinement.

\begin{table}[h]
\centering
\caption{Comparison of reward-model variants on online evaluations}
\label{tab:reward_results}
\begin{tabular}{lc}
\toprule
Model & Macro F1  \\
\midrule
Vanilla GPRO & 77.04  \\
 \; + Scalar Reward Model & 77.31 (+0.27)\\
 \; + Generative Reward Model & 77.66 (+0.62)  \\
\bottomrule
\end{tabular}
\end{table}

\section{Training Efficiency of SRPO}
\label{app:computaion_overhead}
The introduction of the Step-wise Reward Model (SRM) inevitably increases GPU memory consumption and training time. However, since data quality is more critical than sheer volume in Reinforcement Learning (RL), we focus on synthesizing a high-quality dataset. By maintaining a relatively small data scale, we are able to perform iterations with minimal temporal costs.Consequently, while SRPO (Step-wise Reward Policy Optimization) incurs a slightly higher total training cost (approximately $1.5\times$ compared to standard baselines), the training process typically concludes within 1–2 days. This efficiency aligns with common industry practices. 
Finaly, we achieved a Model FLOPs Utilization (MFU) of 35\% during the training phase, demonstrating optimized compute throughput.

\section{Reward Coefficient Settings}
The policy model outputs a sequence of five steps. For steps 1–4, a correct response yields a reward of 0.2 and an incorrect response yields 0. For step 5, a correct outcome yields a reward of 1.0 and an incorrect outcome yields 0. The total reward for an episode is the sum of the rewards across all five steps.

\section{Examples of Generative Stepwise Reward Model Outputs}
Figure \ref{fig:rm_output_examples} shows examples of the reward model's judgments. For each intermediate step of the policy model, the reward model analyzes whether it is correct, provides a plausible justification, and outputs a binary verdict (correct or incorrect).

\section{Quality Results for SRPO}
\begin{figure*}
\centering
\includegraphics[width=0.75\linewidth]{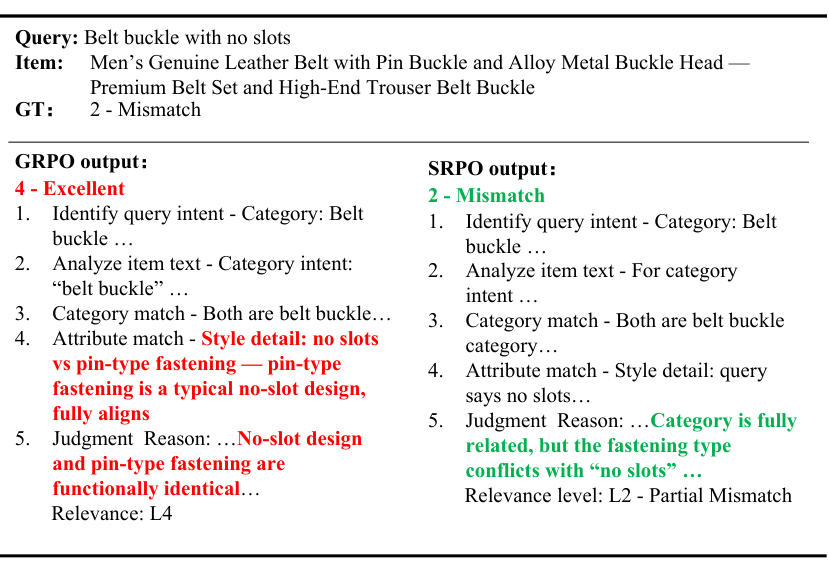}
\caption{Quality results for SRPO.}
\label{fig:policy_qua_result2}
\end{figure*}

\begin{figure*}
\centering
\includegraphics[width=0.75\linewidth]{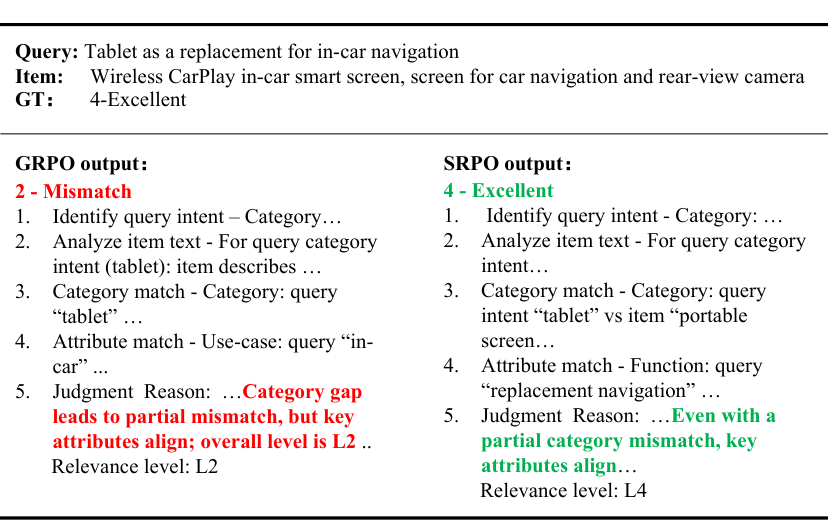}
\caption{Quality results for SRPO.}
\label{fig:policy_qua_result1}
\end{figure*}

\begin{figure*}
\centering
\includegraphics[width=0.75\linewidth]{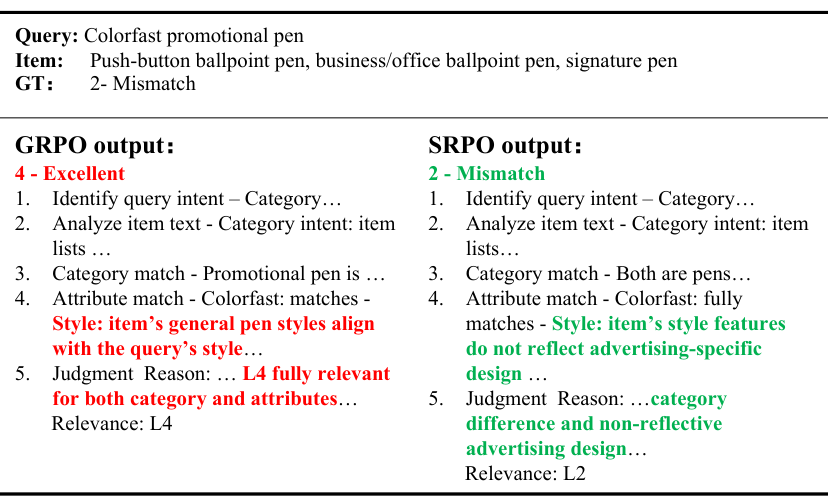}
\caption{Quality results for SRPO.}
\label{fig:policy_qua_result3}
\end{figure*}

Figures \ref{fig:policy_qua_result2}, \ref{fig:policy_qua_result1} and \ref{fig:policy_qua_result3} illustrate the quality results for SRPO. Compared with vanilla GRPO, SRPO demonstrates consistent correct behavior and robust performance across the evaluated scenarios.

\end{document}